\documentclass{elsarticle}

\usepackage{soul}

\usepackage[normalem]{ulem}
\usepackage[numbers]{natbib}

\usepackage{array, booktabs} 
\usepackage[table]{xcolor} 

\usepackage[utf8]{inputenc}
\usepackage[T1]{fontenc}    
\usepackage{hyperref}

\usepackage{amsmath, amsthm}
\newtheorem*{theorem}{Theorem}

\newtheorem*{definition}{Definition}
\usepackage{amsfonts}
\usepackage{amssymb}

\usepackage{graphicx}
\usepackage{xcolor}
\usepackage{geometry}
\usepackage{float}
\usepackage{subcaption}
\usepackage{multirow}
\usepackage{tikz, pgfplots}
\usetikzlibrary{shapes,arrows}
\usetikzlibrary{math}

\usepackage{dsfont}
\usepackage{mathrsfs}

\usepackage{pdflscape}
\usepackage[multiple]{footmisc}

\usepackage{url}            

\usepackage{setspace}
\usepackage[ruled,vlined]{algorithm2e}
\usepackage{algorithmic}

\newcommand{\bsu}{{\boldsymbol{u}}}

\newcommand{\bsy}{{\boldsymbol{y}}}

\newcommand{\RR}{{\mathbb{R}}} 
\newcommand{\EE}{{\mathbb{E}}}

\DeclareSymbolFont{bbold}{U}{bbold}{m}{n}
\DeclareSymbolFontAlphabet{\mathbbold}{bbold}

\newcommand{\calN}{{\mathcal{N}}}

\newcommand{\calX}{{\mathcal{X}}}
\newcommand{\calY}{{\mathcal{Y}}}

\usepackage{mathtools}
\DeclarePairedDelimiterX{\infdivx}[2]{(}{)}{%
  #1\;\delimsize\|\;#2%
}

\newif\ifnotes\notestrue

\def\mfabian#1{{\color{black} #1}}
\def\hfabian#1{}

\def\hpascal#1{}

\def\hcinzia#1{}

\def\hjavier#1{}

\def\hmohammad#1{}

\begin{document}

%

\begin{frontmatter}

\author[1,2]{Pascal Jutras-Dubé}\ead{pascal.jutras-dube@umontreal.ca}
\author[3]{Mohammad B. Al-Khasawneh}\ead{mbmk2020@umd.edu}
\author[3]{Zhichao Yang}\ead{zcyang97@umd.edu}
\author[4]{Javier Bas}\ead{javier.bas@uam.es}
\author[1,2]{Fabian Bastin}\ead{bastin@iro.umontreal.ca}
\author[3]{Cinzia Cirillo\corref{cor1}}\ead{ccirillo@umd.edu}

\affiliation[1]{
organization={Department of Computer Sience and Operations Research, Université de Montréal},
addressline={CP 6128, Succ. Centre-Ville},
city={Montreal, QC},
postcode={H3C 3J7},
country={Canada}}
\affiliation[2]{
organization={CIRRELT, Université de Montréal},
addressline={CP 6128, Succ. Centre-Ville},
city={Montreal, QC},
postcode={H3C 3J7},
country={Canada}}
\affiliation[3]{
	organization={Department of Civil and Environmental Engineering, University of Maryland},
	city={College Park, MD},
	postcode={20742},
	country={USA}}
\affiliation[4]{
organization={Department of Quantitative Economics, Universidad Autónoma de Madrid},
city={Madrid},
postcode={28049},
country={Spain}}

\cortext[cor1]{Corresponding author}

\title{Copula-based transferable models for synthetic population generation}

\begin{abstract}
Population synthesis involves generating synthetic yet realistic representations of a target population of micro-agents for behavioral modeling and simulation. Traditional methods, often reliant on target population samples, such as census data or travel surveys, face limitations due to high costs and small sample sizes, particularly at smaller geographical scales. We propose a novel framework based on copulas to generate synthetic data for target populations where only empirical marginal distributions are known. This method utilizes samples from different populations with similar marginal dependencies, introduces a spatial component into population synthesis, and considers various information sources for more realistic generators. Concretely, the process involves normalizing the data and treating it as realizations of a given copula, and then training a generative model before incorporating the information on the marginals of the target population. Utilizing American Community Survey data, we assess our framework’s performance through standardized root mean squared error (SRMSE) and so-called sampled zeros. We focus on its capacity to transfer a model learned from one population to another. Our experiments include transfer tests between regions at the same geographical level as well as to lower geographical levels, hence evaluating the framework’s adaptability in varied spatial contexts. We compare Bayesian Networks, Variational Autoencoders, and Generative Adversarial Networks, both individually and combined with our copula framework. Results show that the copula enhances machine learning methods in matching the marginals of the reference data. Furthermore, it consistently surpasses Iterative Proportional Fitting in terms of SRMSE in the transferability experiments, while introducing unique observations not found in the original training sample.

\end{abstract}
\end{frontmatter}

\section{Introduction}

Population synthesis refers to models that aim at constructing artificial datasets whose characteristics replicate that of a population of interest.
Population synthesizers can produce a set of agents with detailed sociodemographic and socioeconomic information at both the individual and household levels, maintaining the structural coherence of the population they emulate.
This capability is particularly valuable in agent-based models, such as transportation models based on micro-simulation, where understanding the spatial implications of policies is critical \cite{ArenTimmHofm07,PritMill12, MullAxha11,GuoBhat14}.  

In order to properly reflect the population properties it aims to mimic, a synthetic population must share the same joint distribution of the variables that it encapsulates. 
In the particular, in the case microsimulation of travel behavior, this property must hold for the geographic area of interest. Various methods have been proposed to achieve this goal but, with a few notable exceptions ~\cite{BartToin13} most of them rely on samples from a target population, such as census data or travel survey, which can be costly to obtain. 
This often result in limited sample sizes, especially at smaller geographical scales.
Additionally, some regions, such as those at the census tract level, may lack detailed data entirely, providing only marginal totals--which effectively represent the attribute distributions of these regions. 
Traditional methods like re-weighting estimate sampling weights from the attribute distributions in each district and simulate the population from the weighted samples \cite{Castiglione2014, Rich2018, HorlBala21}. 
However, re-weighting cannot produce attribute combinations not observed in the training samples but present in the actual population. 
More recent approaches, based on generative models, can generate new, out-of-sample attribute combinations due to their probabilistic nature but do not explicitly integrate the attribute distributions of the area under study \cite{SunErat15, BoryRichPere19, kim2023deep}. 
Consequently, there is an increasing demand for population synthesis methods that can generate new samples while accurately matching the aggregate totals of the studied region. These methods should leverage insights from data-rich environments to effectively synthesize populations in data-scarce areas, thereby enhancing the transferability of models across different geographical contexts.

To address this gap, we introduce a framework capable of generating synthetic data for a target population by utilizing only known marginal totals, in combination with a sample from another population that exhibits similar structural relationships among variables--such as a population from an adjacent region or a larger geographic area. 
Our approach integrates copula theory with machine learning (ML) generative modeling techniques to separate the learning of dependency structures from that of marginal distributions. 
This separation facilitates the framework's application across different populations with varying marginal distributions. 
The generative model focuses on capturing the structural relationships between variables from the available sample, which is essential for accurately fitting multivariate dependencies. 
Meanwhile, copula normalization ensures that the marginal distributions of the synthesized data align with those of the reference data. 
We evaluate several methods including Bayesian Networks, Variational Autoencoders, and Generative Adversarial Networks both alone and in combination with copulas. 
Our findings, based on data from the American Community Survey (ACS), demonstrate that versions incorporating our copula framework more effectively match the marginal totals of the reference data while also retaining the capability to generate previously unseen examples. 
We analyze this model's transferability across various geographical levels, including state, county, Public Use Micro Areas (PUMA), and census tract.

A key advantage of our approach is that it eliminates the need to select a specific copula family~\cite{DuraSemp16}.
This flexibility allows to choose the generative model that best fits the context of the population under study.
It also gives the opportunity to fully leverage the capabilities of probabilistic generative models, which have been shown in recent literature to be highly effective in capturing complex dependencies between variables and generating diverse data \cite{SunErat15, BoryRichPere19, kim2023deep}.

The rest of the paper is structured as follows: Section \ref{sec:review} reviews population synthesis literature, focusing on recent methodologies. 
Section \ref{sec:method} presents our method and the specific generative models used in our experiments.
Section \ref{sec:exp} is devoted to the details of our experimental setup and the results obtained, illustrating the versatility of our framework across different generative models and its effectiveness in model transferability at various geographical and sub-geographical levels. 
Lastly, conclusions and potential avenues for future research are discussed in Section \ref{sec:conclusion}.

\section{Literature review}\label{sec:review}

The purpose of a population synthesizer is to create an artificial, yet realistic, population of obervations from existing, but limited, disaggregated datasets. The information generated can be used for varios purposes, including modeling, optimization, simulation, or, in general, generating new information for an application of interest~\cite{ChapTailDrog22}. In transportation research, these synthesizers have been extensively employed~\cite{YameGastHankVand21}. Particularly,\citet{ElurPinjGuoSeneSrinCoppBhat08}, who developed CEMUS, a microsimulator that uses aggregated socioeconomic data to create detailed synthetic datasets for individual-level activity-travel pattern modeling. Similarly, \citet{Brad10} used synthetic populations for forecasting models in urban planning, while \citet{AuldJavaMoha12} and \citet{ZiemJoubNage18} applied them in dynamic simulators for route choice generation and accessibility measure computation, respectively.

A desirable property of any population synthesizer is to preserve the general characteristics of the population, which leads to different technical approaches.
The most widely used, until recently, has probably been the Iterative Proportional Fitting (IPF), initially proposed by \citet{DemiStep40}, and popularized in transportation by~\citet{DuguJungMcFa76}.
IPF selects households from the source sample trying to match some given marginals totals, requiring a fitting stage and an allocation stage.
In the fitting step, a contingency table is computed from the seed table (the source sample) and the marginals totals.
In the allocation phase, households are randomly selected from the seed table to match the frequency given in the contingency table.
There is an abundant literature on empirical applications of IPF---reviewed, for instance, by \citet{MullAxha11}, as well as on its technical aspects~\citep{ArenTimm04,SalvMill05,AuldWies09,YeKondPendSanaWadd09,GuoBhat14}.
The method nevertheless presents several important flaws. In this regard, \citet{PritMill12} address computational memory restrictions, while \citet{GuoBhat14} explore how to avoid sampling zero issues.
\citet{YeKondPendSanaWadd09} consider simultaneously fitting different types of agents, proposing a heuristic approach called Iterative Proportional Updating (IPU) to overcome the disadvantages of the standard IPF. However, they fail to accommodate the new synthetic information at multiple geographical resolutions simultaneously, leading to a loss of representativeness.
\citet{KondYouGariPend16} extend their efforts, proposing an enhanced IPU algorithm that accounts for constraints at different levels of spatial resolution when generating a synthetic population.

However, as \citet{FaroBierHurtFlot13} point out, fitting a contingency table to the available data may entail errors if the information is not complete or has been manipulated. 
In fact, these potential errors cannot be contrasted due to absence of a complete real data set. 
Thus, these authors propose a Markov Chain Monte Carlo (MCMC) simulation-based approach that uses partial views of the joint distribution of the real population obtained from the census to draw from it high-dimensional synthetic populations that would otherwise have been impossible to produce using IPF. 
MCMC technique has been further explored by \citet{CasaMullFourEratAcha15} and \citet{SaadMustTellCool16, SaadMustTellFaroCool16}.

Another category of methods for synthesizing populations, the Combinatorial Optimization Methods, proceed differently.
They divide an area into mutually exclusive subareas where a distribution of the attributes of interest is available.
Then a sample taken over the whole population is fitted to the given set of marginals for each subarea.
\citet{huang2001comparison} offer a comparison of these two approaches in the context of the creation of small-area microdata. 
However, the assumptions about the data that these two families of methods require cannot always be met. 
This led \citet{BartToin13} to develop a new type of generator, applied to the case of Belgium. Their technique works in a three-step process. 
First, a pool of individuals pertaining to a certain household is generated; then, the household joint distribution is estimated and stored in a contingency table; finally, the synthetic households are constructed by making random draws from the pool of individuals, preserving the distribution computed  in  the  second  step.
They compare the results of this method with those of the extended IPF described in \citet{GuoBhat14}, observing that the new generator performs better. 
Another effort to offer an alternative approach is due to \citet{sun2018hierarchical}, who propose a hierarchical mixture model to generate representative household structures in population synthesis. 
Their framework comprises a probabilistic tensor factorization, a multilevel latent class model, and a rejection sampling process. 
They test their procedure on the Household Interview Travel Survey data of Singapore, being able to generalize the associations among attributes as well as to reproduce structural relationships among household members.

Yet, it is possible to observe a recent increasing trend in the substitution of iterative fitting algorithms for the generation of synthetic populations by alternative approaches.
One of these new perspectives is the application of Copula Generative Models.
Copulas are a mathematical probability tool that allows the modeling of random variables that have an intrinsic relationship to each other. 
The dependence structure can be decoupled from the original available information so that any set of new synthetic data can be created having that same structure.
We refer to \citet{Gene07} for a gentle introduction on the topic, and to the books by \citet{Joe97, Joe15}, \citet{Nels06}, and \citet{DuraSemp16}, for readers seeking concrete mathematical proofs, theorems and derivations related to copulas.
The selection of a specific copula remains challenging~\cite{KausCiriBast19}, especially in the case of discrete copula. 
\citet{AvrChanLecu2009} study the NORTA correlation-matching problem for modeling dependence of discrete multivariate distributions, while \citet{Niko13} reviews copula-based approaches with discrete multivariate response data.
Copulas have been successfully applied in various domains, including finance and actuarial sciences~\cite{CherLuciVecc04, JawoDuraHard13}, transportation~\cite{BhatElur09,PinjBhatHens09,RamaSikdPinj10,KaoKimLiuCuiBhad12,BornYasmYouElurBhatPend14}, water resources management~\cite{BorgPfluHallHoch15}, and simulation modeling of arrival rates in call centers~\cite{OresReegLecu2016, JaouLecuDelo2013}.

Bayesian network (BN) is another alternative approach that has recently gained momentum as a tool to perform population synthesis. 
This method encodes the dependency relationships among predictors using a graphical model in which nodes represent variables, links capture their conditional dependencies, and probability distributions are assigned to each node, conditional on its parents. 
As expressed in the seminal work by~\citet{SunErat15}, if the conditional structure of the data generating process is known, inference can be based on the underlying probabilities
and the population can then be generated accordingly by sampling from the joint distribution. Since this information is typically not available, the authors propose that the graph structure of the data may be learned through a scoring approach. More recent examples of the use of BN for population synthesis are given by~\citet{ZhanCaoFeygTangShenPozd19}, who make use of traditional survey data and digital records of networking and human behavior to generate connected synthetic populations using BN; ~\citet{HorlBala21}, from their part, generate a synthetic travel demand based on open data.

Other approaches to the generation of synthetic populations fall more along the lines of ML.
The most popular of these methodologies are probably generative models~\citep{Bish06}, given their success in generating artificial images.
A generative model captures the dependency structure of the variables of a dataset and trains a generator to reproduce samples preserving their joint distribution.
Among generative models, two families stand out: Variational Autoencoders
(VAE)~\citep{KingWell14} and Generative Adversarial Networks (GAN)~\citep{GoodPougMirzXuWardOzaiCourBeng14}. 
\citet{BoryRich21} use a conditional VAE to study travel preference dynamics using a synthetic pseudo-panel approach, while \citet{boquet2020variational} propose a VAE model to generate traffic data. 
\citet{YazdPattFaroo21} use a GAN to infer travel mode from GPS trajectory data and derive trip information from travel survey.
The work of~\citet{GuntSimpRogg20} delves along the same lines.
On the contrary, \citet{YinSheeFeygPaiePozd18} focus on the use of deep generative models to generate synthetic travelers’ mobility patterns that replicate the statistical properties of a sample of actual travelers.
\citet{kim2023deep} apply regularization techniques to ML generative models to enhance the generation of realistic agents. 
Their approach focuses on maximizing the production of realistic combinations, which exist in the actual population but are not necessarily present in the training sample, while simultaneously minimizing the generation of infeasible combinations, unobservable in the population.
Conditional tabular GAN (CTGAN) and tabular VAE (TVAE)~\citep{XuSkouCuesVeer19} are adaptations of regular GAN and VAE to tabular data with a mix of potentially highly imbalanced discrete columns and multimodal non-Gaussian-like continuous columns.  

In this paper, we integrate copula theory with ML generative models, harnessing the strengths of both approaches. 
Our use of ML generative models aims to leverage their robust capacity for generalization and their ability to generate realistic yet diverse combinations. 
By incorporating copula theory, we effectively decouple the modeling of variable dependency structures from their marginal distributions. 
This decoupling is crucial as it allows us to integrate the marginal distribution information of the target population. 
Consequently, this integration enables the transferability of the model across different populations, a key aspect in population synthesis.
In the next section, we describe the generative models that we use for our experiments (BN, CTGAN, and TVAE), and we detail how we leverage copula theory to enable model transferability.

\section{Methodology}
\label{sec:method}

Consider a sample of size $N$ from a random vector $X$, from which we want to generate a synthetic population of arbitrary size.
As highlighted in the previous section, many techniques can be used to generate a synthetic population, yet we will only select, in addition to IPF, the most promising approaches. We first briefly introduce Bayesian Networks, Conditional Tabular Generative Adversarial Network (CTGAN), and Tabular Variational Autoencoders (TVAE), that will serve as our benchmark methods.
We then discuss copulas and their link to normalization procedures in ML, and conclude with their use in the generation of synthetic populations.

\subsection{Bayesian Networks}
\label{sec:bn}

A Bayesian Network (BN) is a probabilistic model that represents a set of random variables $X_i$, $i = 1,\ldots,d$, and their conditional dependencies in the form of a directed acyclic graph $G$.  
Each node corresponds to one random variable, and the arcs express the conditional dependencies between them. If no path exists between two nodes, the corresponding random variables are independent. 
The joint probability distribution of $X$ can be decomposed as the product of the marginal distributions of $X_i$, conditioned on its parents $\Pi_i$, $i = 1,\ldots,d$, using the chain rule 
$$
P(X\mid G, \Theta) = \prod_{i=1}^d P(X_i\mid \Pi_i, \theta_i),
$$
where $\theta_i$ are the parameters of the distribution of $X_i$ and $\Theta = \left\lbrace \theta_1,\ldots,\theta_d \right\rbrace$.

\begin{figure}[htbp]
	\centering
	\includegraphics[width=0.6\textwidth]{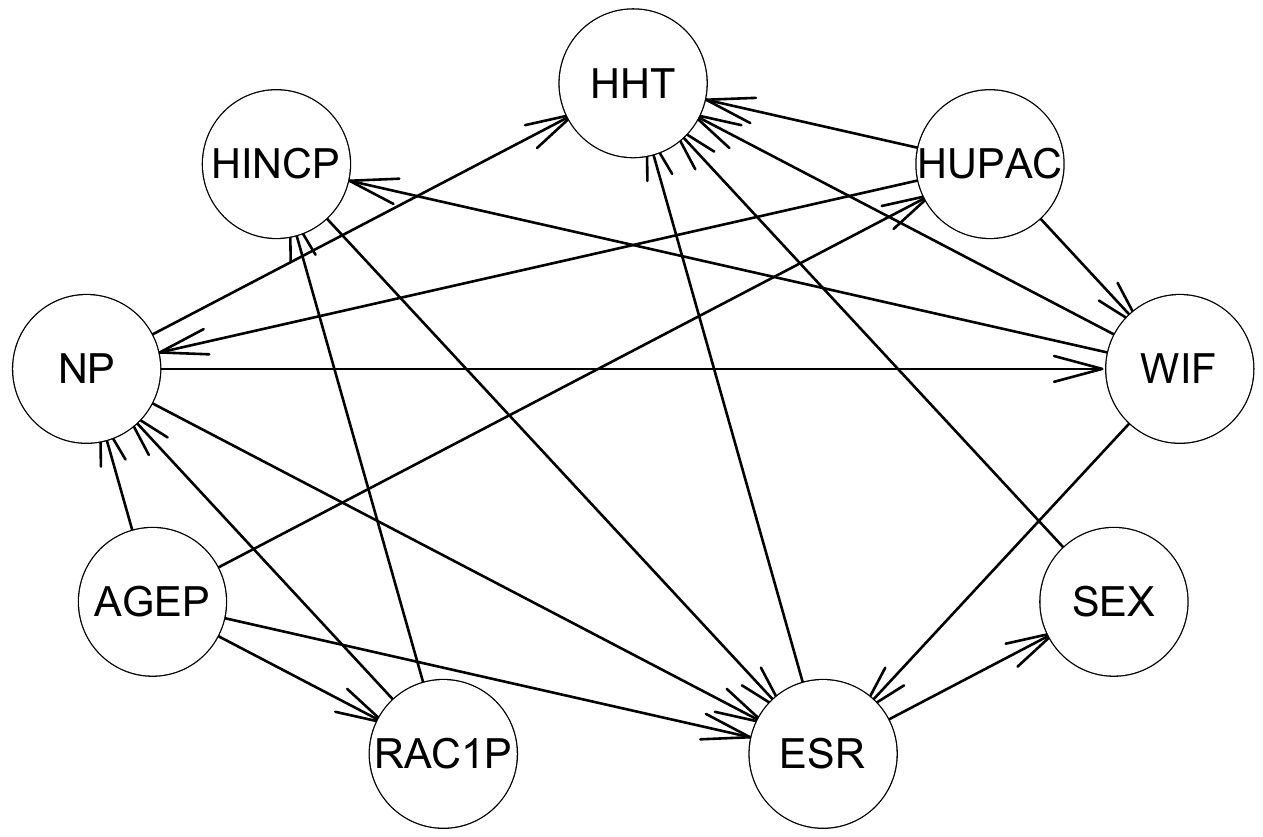}
	\caption{Example of a Bayesian network found by the greedy search algorithm.}
	\label{fig:bn} 
\end{figure}

Both the graph structure $G$ and the parameters $\Theta$ are learned. 
This process is often referred to as \textit{structural learning}
and has two stages: model selection and model optimization. In the selection stage, a score function is used to quantify how well a hypothetical graph structure fits the data.
The minimum description length (MDL) is a popular scoring function consisting of two components that estimate the structural complexity and the likelihood of the data given the model~\citep{LamBacc94}. 
For discrete BN, the variables conditioned on their parents are assumed to follow multinomial distributions and their parameters are estimated by maximum likelihood estimation. 
The goal of the model optimization stage is to identify the hypothetical structure with the highest score. 
However, the evaluation of all possible graph structures has an impractically high time complexity~\citep{Coop90}.
Instead, it is possible to apply a heuristic search that greedily chooses a topological ordering of the variables, and optimally identifies the best parents for each variable given this ordering~\citep{HeckGeigChick95}, as illustrated in Figure~\ref{fig:bn}, depicting a Bayesian Network built on data from the ACS dataset.
 
Once the structural learning is completed, a synthetic population can be drawn from the the factorized joint probability distribution defined by the Bayesian network.
More precisely, we generate values via rejection sampling, and refer
the reader to \citet{SunErat15} for more details on the application of BN in population synthesis.

\subsection{Conditional Tabular Generative Adversarial Network}

Generative Adversarial Networks (GANs)~\citep{GoodPougMirzXuWardOzaiCourBeng14} are a class of generative deep learning models trained adverserially. In GANs, a generative model competes against a discriminative model, which learns to distinguish whether a sample is from the generative model’s distribution or the actual data distribution. This ongoing competition drives both models to continually improve. As a result, the generative model eventually produces synthetic data that is virtually indistinguishable from real data.

The generator, denoted a $G_{\theta}$, is a neural network characterized by parameters $\theta$. 
It aims to approximate the data distribution $p_X$ by generating synthetic samples $G_{\theta}(Z)$ from a prior random distribution $Z \sim p_Z$.
On the other hand, the discriminator $D_\phi$ is a neural network with parameters $\phi$. 
Its scalar output $D_\phi(x)$ indicates the likelihood that a sample $x$ is derived from the actual data rather than from the generator.
The discriminator $D_\phi$ is trained to optimize the probability of accurately determining whether a sample was drawn from the training set or generated by $G_\theta$.
Concurrently, $G_\theta$ is trained to reduce the likelihood that $D_\phi$ correctly classifies its output. The objective function can be expressed as:
$$
\min_{\theta}\max_{\phi} \EE_{X\sim p_X}[\log D_\phi(X)] + \EE_{Z\sim p_{Z}}[\log(1 - D_\phi(G_\theta(Z)))].
$$

While GANs have demonstrated remarkable capabilities in producing high-quality synthetic images, they weren't initially conceived for handling tabular data. 
This gap inspired the creation of Conditional Tabular GANs (CTGAN). 
CTGANs are adept at managing tabular data, accommodating both potentially highly imbalanced discrete columns and multimodal non-normal continuous columns.

CTGAN incorporates a conditional generator within the GAN architecture to address class imbalances in categorical columns. This integration is achieved through three components: the conditional vector, the generator loss, and the training-by-sampling technique. The conditional vector indicates whether a specific discrete column, $D_{i^*}$, should have a particular value~$k^*$. 
Given that $D_{i^*}$ is depicted as a one-hot encoded vector $d_{i^*}$, this condition can be expressed as $d_{i^*}^{(k^*)} = 1$. 
Every discrete column $D_i$ has a corresponding mask vector $m_i$, mirroring its one-hot encoding $d_i$, to represent the condition as:
$$
m_i^{(k)} = 
\begin{cases}
1&\text{if }i = i^* \text{ and } k = k^*\\
0&\text{otherwise}.
\end{cases}
$$
The conditional vector emerges from the concatenation of these masks vectors. 

During the feed-forward pass, no inherent mechanism encourages the conditional generator to yield the anticipated output $D_{i^*}=k^*$.
To enforce it, the generator's loss is augmented by integrating the cross-entropy between the generated discrete columns and the conditional vector.
The training-by-sampling strategy samples both the conditional vector and training data, guiding the model to evenly explore all possible values within discrete columns.
This method begins by arbitrarily selecting a discrete column, constructs a probability mass function spanning the value spectrum of the chosen column, and then samples a value based on this function.
It concludes by constructing the conditional vector that symbolizes the selected column and its value. 
For a more comprehensive understanding of CTGAN's mechanics, we direct readers to \citet{XuSkouCuesVeer19}.

\subsection{Tabular Variational Autoencoder}

\begin{figure}[htbp]
    \centering
    \includegraphics[width=0.5\textwidth]{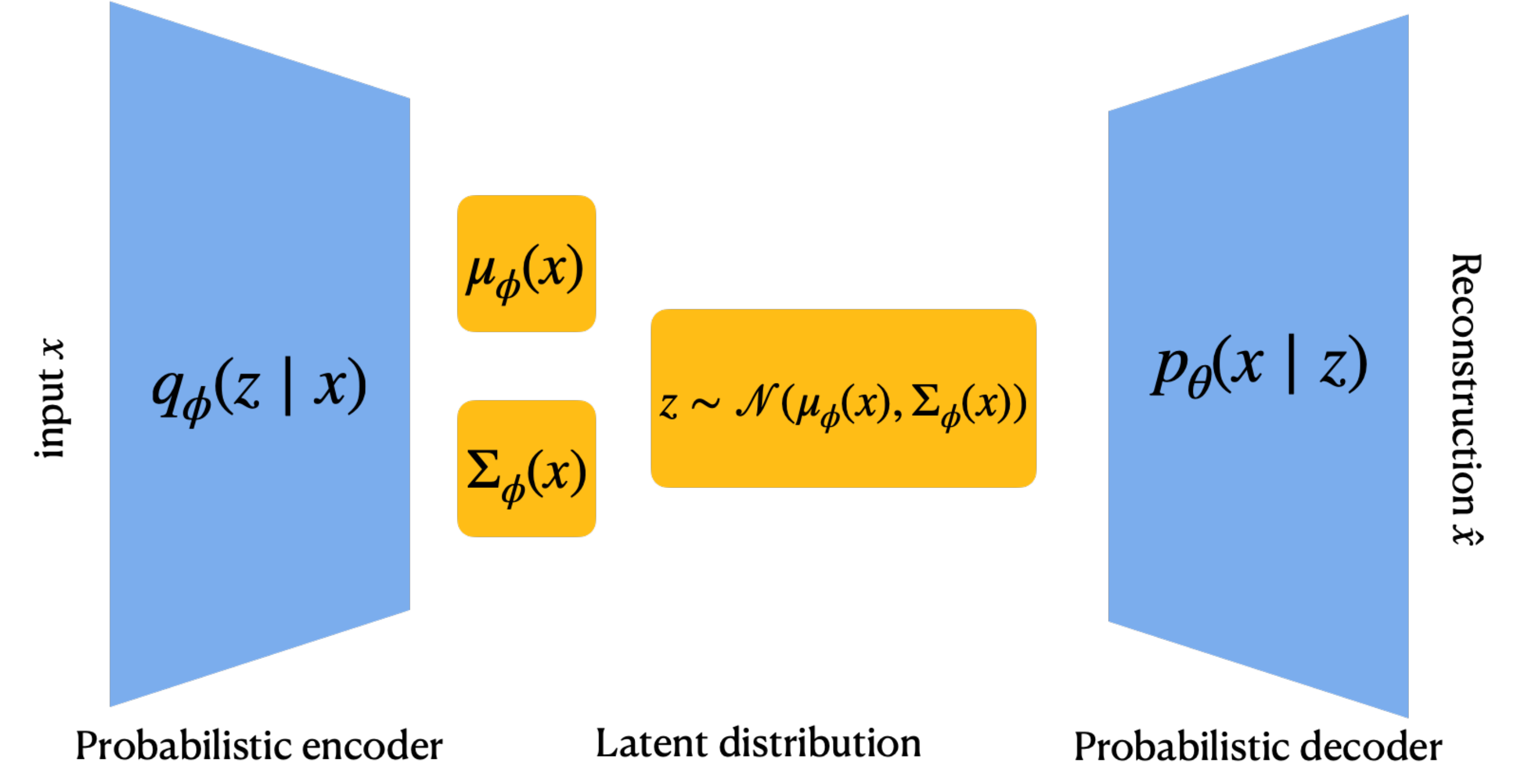}
    \caption{Schematic description of a variational autoencoder for which $q_\phi(\cdot\mid x)$ is assumed to be the family of Normal distributions.}
    \label{fig:vae} 
\end{figure}

The architecture of a Variational Autoencoder (VAE)~\citep{KingWell14}, considered by \citet{BoryRichPere19} in the context of synthetic populations, is analogous to an autoencoder.
A probabilistic encoder $q_\phi(Z\mid X)$ is meant to map the input to a multivariate latent distribution and a probabilistic decoder $p_\theta(X\mid Z)$ maps back the latent distribution to the data space. 
The probabilistic encoder and decoder are represented by neural networks, with parameters $\phi$ and $\theta$, respectively. One would like to approximate the data distribution by first sampling from a known prior distribution. Thus, let $z$ represent a latent encoding of a data point $x$ and let's consider their joint distribution $p_\theta(X,Z)$.
Conditioning on $Z$ gives us the parameterized approximation $p_\theta(X)$ of the data distribution is obtained: 
$$
p_\theta(x)=\int_z p_\theta(x\mid z)p_\theta(z)dz.
$$
However, the computation of $p_\theta(x)$ is usually intractable.
For this reason, it is necessary to introduce an auxiliary parameterized family of functions, typically the Normal distributions $\calN(\mu(x),\Sigma(x))$, to approximate the posterior distribution;
$$
q_\phi(z\mid x) \approx p_\theta(z\mid x), %
$$
as schematized in Figure~\ref{fig:vae}. 
The objective is to make $q_\phi(z\mid x)$ as close as possible to $p_\theta(z\mid x)$ by minimizing a distance measure between the distributions. 
In addition, it is desirable to optimize the generative model to reduce the reconstruction error between the input and output. In this regard, maximizing the evidence lower bound (ELBO), defined as 
\begin{align*}
	L(x) 
	&= \EE_{z\sim q_\phi(\cdot\mid x)}\left[\frac{p_\theta(x, z)}{q(z\mid x)}\right]\\
	&= \ln p_\theta(x) - D_{KL}\left(q(z\mid x) \parallel p_\theta(z\mid x)\right)\mfabian{,}
\end{align*}
is equivalent to minimizing the reverse Kullback-Leibler divergence between the approximation and true posteriors and jointly maximizing the log-likelihood of the data. This is typically achieved by using a stochastic gradient descent procedure. 
For more details, we refer the reader to \cite{KingWell14}. Nevertheless, it is worth mentioning at this point that A Tabular Variational Autoencoder (TVAE) adapts a VAE to tabular data with the same preprocessing and loss function modification procedures as in the CTGAN framework.

\subsection{Copula-based generation}

The casuistry one is confronted with in relation to the training sample and the marginals of the target population is varied. Namely, most household surveys and similar studies rely on detailed but small samples, while more comprehensive information about specific household characteristics is often available elsewhere. Typically, data from these surveys are presented as multivariate vectors, implicitly capturing the dependencies between their components. Nevertheless, we sometimes do have access to the distributions of individual marginals at the population level. In some cases, we want to focus on a sub-area covered by the sample, resulting in a limited number of observation vectors, while we have comprehensive knowledge of the distribution of individual factors. Additionally, there are situations where we only have access to marginal distributions and no detailed household vector data. For example, in the context of generating census data, it is reasonable to conjecture that marginals encode demographic elements, while multidimensional dependencies capture more complex socioeconomic patterns shared across related regions. Assuming these dependencies can be adequately captured by a sample from a different geographical sector, our aim is to exploit this dependency structure in the area of interest. We thus need a tool to separate the dependencies structure to the specific marginal distributions, which is provided by the copula theory.

A $d$-dimensional copula $C$ is a multivariate cumulative distribution function (CDF) on $[0, 1]^d$ having all marginals uniformly distributed on $[0, 1]$ \citep{DuraSemp16, Joe97, Joe15, Nels06, OkhrRistXu17}.
A fundamental result, due to \citet{Skla59}, states that any multivariate distribution can be represented by means of a copula.

\begin{theorem}[Sklar's theorem]
	Let $H$ be a multivariate distribution function with margins $F_1,\ldots, F_d$, then there exists a copula $C$ such that
	\begin{equation}
		H(x_1, \ldots, x_d) = C \left( F_1(x_1), \ldots, F_d(x_d) \right), \quad x_1, \ldots, x_d \in \overline{\RR}.
		\label{eq:Copulas}
	\end{equation}
	If $F_\mfabian{i}$, $i = 1, \ldots, d$, are continuous then $C$ is unique. Otherwise $C$ is uniquely determined on the Cartesian product of the range of the marginals $F_1 \left( \overline{\RR} \right) \times \ldots \times F_d \left( \overline{\RR} \right)$. Conversely, if $C$ is a copula and $F_1, \ldots, F_d$ are univariate distribution functions, then the function $H$ defined above is a multivariate distribution function with margins $F_1, \ldots, F_d$.
\end{theorem}

Let $X$ be the random vector behind the population $\calX$, from which we extract a sample, while we are interested in the population $\calY$, governed by the random vector $Y$.
We assume from now that $X$ and $Y$ share the copula $C$.
This assumption is valid when $\calX$ is a Monte Carlo sample from $\calY$, and we argue in the following that it also holds when one population is a subset from the second one or when they are two different populations with similar characteristics.
We formalize this idea by introducing the definition of a shared copula.

\begin{definition}[Shared copula]
\label{def:shared_copula}
Let $F^X$ and $F^Y$ be multivariate distribution functions of random vectors $X$ and~$Y$, respectively, and denote their margins by $F^X_1,\ldots, F^X_d$ and $F^Y_1,\ldots, F^Y_d$.
The random vectors $X$ and~$Y$ are said to share a copula $C$ if
there exists a copula $C$ such that for any vector $(x_1, \ldots, x_d) \in \overline{\RR}^d$, $F^X(x_1, \ldots, x_d) = C \left( F^X_1(x_1), \ldots, F^X_d(x_d) \right)$ and $F^Y(x_1, \ldots, x_d) = C \left( F^Y_1(x_1), \ldots, F^Y_d(x_d) \right)$.
\end{definition}

Since a copula is a multivariate distribution on $[0,1]^d$, the first step is to cast the observations as vectors in $[0,1]^d$.
This can be easily achieved, as it is standard in machine learning to apply some normalization procedure to the data (see for instance~\citet{LaroLaro14}), by considering the empirical CDF (ECDF) of each feature $X_i$, $i = 1,\ldots,d$, defined as
\begin{equation}
\hat{F}_i(x) = \frac{1}{\mfabian{m_i}}\sum_{j=1}^{\mfabian{m_i}}
\mathds{1}\mfabian{\left(x_i^{(j)} \leq x\right)},
\label{eq:ecdf}
\end{equation}
where $\mathds{1}$ is the indicator function and
$\left\{ x_i^{(1)},\ldots,x_i^{(m_i)} \right\}$ is the ordered set of distinct observations (i.e. $x_i^{(p)} < x_i^{(q)}$ if $p < q$) for the $i$-th marginal.
We refer to this normalization process as `copula normalization'.
The multivariate distribution underlying the normalized observations can then be estimated using any of the previously considered methods, without the need to specify a copula family.

Suppose now we wish to generate synthetic data of a target population from which we only have information on the marginals, and that we have access to a sample of another source, sharing the structure of the target.
Our population synthesis procedure is summarized in Algorithm~\ref{algo:generation}.
\begin{algorithm}[htbp]
\begin{enumerate}
    \item[Step 1] Normalize the source population data using the ECDFs $\hat{F}_i(\cdot)$, $i = 1,\ldots,d$. 
    \item[Step 2] Train the model on the normalized data to learn a copula $C$.
    \item[Step 3] Generate a synthetic population of vectors in $[0,1]^d$ by sampling from $C$.
    \item[Step 4] Transform any generated vector $\bsu = (u_1,\ldots,u_d)$ in a vector $\bsy$ in the target population as
    $$
    \bsy = \left( (F_1^Y)^{-1}(u_1),\ldots, (F_d^Y)^{-1}(u_d)\right),
    $$
    where $(F^Y_i)^{-1}(\cdot)$ is the pseudo-inverse distribution function of the $i$-th target marginal, defined as
    $$
    (F^Y_i)^{-1}(u) = \min \left\lbrace \mfabian{x_i^{(j)}}: F^Y_i\left(\mfabian{x_i^{(j)}}\right) \geq u \right\rbrace.
    $$
\end{enumerate}
\caption{Synthetic population generation}
\label{algo:generation}
\end{algorithm}

\begin{figure}[htbp]
\begin{center}
	\begin{subfigure}{0.48\textwidth}
	\begin{tikzpicture}
		
		\begin{axis}[
			anchor=origin,  
			x=4.5cm, y=4.5cm,   
			xmin=0, xmax=1.2,
			ymin=0, ymax=1.2,
			ytick={0,0.2,0.4,0.6,0.8,1.0},
			xticklabels={},
			extra x ticks={0.2,0.5,0.75,1.0},
			extra x tick labels={$x_i^{(1)}$,$x_i^{(2)}$,$x_i^{(3)}$,$x_i^{(4)}$},
			ylabel={$\hat{F}_i(x)$},
			axis y line=left,
			axis x line=left,
			]
			
			\draw[red,fill=red] (axis cs:0.2,0.2) circle (1pt);
			\addplot[red, domain=0:0.2] {0};
			\draw[red,fill=red] (axis cs:0.5,0.4) circle (1pt);
			\addplot[red, domain=0.2:0.5] {0.2};
			\draw[red,fill=red] (axis cs:0.5,0.4) circle (1pt);
			\addplot[red, domain=0.5:0.75] {0.4};
			\draw[red,fill=red] (axis cs:0.75,0.8) circle (1pt);
			\addplot[red, domain=0.75:1.0] {0.8};
			\draw[red,fill=red] (axis cs:1.0,1.0) circle (1pt);
			\addplot[red, domain=1.0:1.1] {1.0};
			
			\addplot[dotted, blue] coordinates {(0.2,0)(0.2,0.2)};
			\addplot[dotted, blue] coordinates {(0.5,0)(0.5,0.4)};
			\addplot[dotted, blue] coordinates {(0.75,0)(0.75,0.8)};
			\addplot[dotted, blue] coordinates {(1.0,0)(1.0,1.0)};
			
			\addplot[dotted, blue] coordinates {(0,0.2)(0.2,0.2)};
			\addplot[dotted, blue] coordinates {(0,0.4)(0.5,0.4)};
			\addplot[dotted, blue] coordinates {(0,0.8)(0.75,0.8)};
			\addplot[dotted, blue] coordinates {(0,1.0)(1.0,1.0)};
			
		\end{axis}
	\end{tikzpicture}
    \caption{Empirical cumulative distribution function.} \label{fig:ECDF}
    \end{subfigure}
\hspace*{\fill}
	\begin{subfigure}{0.48\textwidth}
	\begin{tikzpicture}
		
		\begin{axis}[
			anchor=origin,  
			x=4.5cm, y=4.5cm,   
			xmin=0, xmax=1.2,
			ymin=0, ymax=1.2,
			ytick={0,0.2,0.4,0.6,0.8,1.0},
			xticklabels={},
			extra x ticks={0.2,0.5,0.75,1.0},
			extra x tick labels={$x_i^{(1)}$,$x_i^{(2)}$,$x_i^{(3)}$,$x_i^{(4)}$},
			ylabel={$\tilde{F}_i(x)$},
			axis y line=left,
			axis x line=left,
			]
			
			\draw[red,fill=red] (axis cs:0.2,0.2) circle (1pt);
			\addplot[red, domain=0:0.2] {x};
			\draw[red,fill=red] (axis cs:0.5,0.4) circle (1pt);
			\addplot[red, domain=0.2:0.5] {1/3*0.2+2/3*x};
			\draw[red,fill=red] (axis cs:0.5,0.4) circle (1pt);
			\addplot[red, domain=0.5:0.75] {1.6*x-0.4};
			\draw[red,fill=red] (axis cs:0.75,0.8) circle (1pt);
			\addplot[red, domain=0.75:1] {0.8*x+0.2};
			\draw[red,fill=red] (axis cs:1.0,1.0) circle (1pt);
			\addplot[red, domain=1.0:1.1] {1.0};

			\addplot[dotted, blue] coordinates {(0.2,0)(0.2,0.2)};
			\addplot[dotted, blue] coordinates {(0.5,0)(0.5,0.4)};
			\addplot[dotted, blue] coordinates {(0.75,0)(0.75,0.8)};
			\addplot[dotted, blue] coordinates {(1.0,0)(1.0,1.0)};
			
			\addplot[dotted, blue] coordinates {(0,0.2)(0.2,0.2)};
			\addplot[dotted, blue] coordinates {(0,0.4)(0.5,0.4)};
			\addplot[dotted, blue] coordinates {(0,0.8)(0.75,0.8)};
			\addplot[dotted, blue] coordinates {(0,1.0)(1.0,1.0)};
		\end{axis}
	\end{tikzpicture}
    \caption{Empirical cumulative distribution function with piecewise linear interpolations.} \label{fig:RECDF}
    \end{subfigure}
\end{center}
\caption{\mfabian{Continuous approximation of an empirical cumulative distribution function.}}
\end{figure}
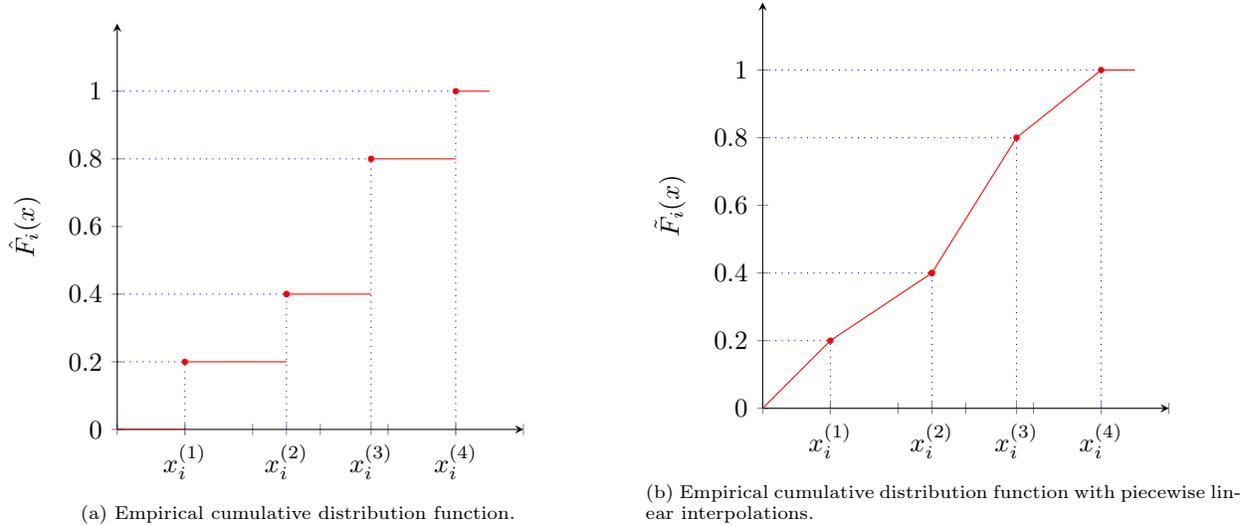

Algorithm~\ref{algo:generation} implicitly assumes that the marginal distribution functions of the source and target populations have the same ranges. This assumption is often violated, especially when the marginals follow discrete distributions, which is the case when we use the ECDFs \eqref{eq:ecdf}, as illustrated in Figure~\ref{fig:ECDF}.
We address this issue by relaxing any discrete distribution function as a continuous distribution with the following heuristic.
We construct the relaxed ECDF $\tilde{F}_i(\cdot)$ as the continuous piecewise linear function obtained by considering the linear interpolations between consecutive values $\hat{F_i}\left(x_i^{(k)}\right)$ and $\hat{F_i}\left(x_i^{(k+1)}\right)$, $k = 1,2,\ldots,m_i-1$, as depicted in Figure~\ref{fig:RECDF}.
The relaxed marginals being now continuous, each one is uniformly distributed on $[0,1]$.
We now extend the copula $C$ to the domain $[0,1]^d$ by setting $C\left(\tilde{F}_1(x_1),\ldots,\tilde{F}_d(x_d)\right)$
as the linear interpolation, component by component, from $C\left(\hat{F}_1\left(x_1^{(k_1)}\right),\ldots,\hat{F}_i\left(x_i^{(k_i)}\right),\ldots,\hat{F}_d\left(x_d^{(k_d)}\right)\right)$ to $C\left(\hat{F}_1\left(x_1^{(k_1)}\right),\ldots,\hat{F}_i\left(x_i^{(k_i+1)}\right),\ldots,\hat{F}_d\left(x_d^{(k_d)}\right)\right)$,
with $x_i^{(k_i)} \leq x_i \leq x_i^{(k_i+1)}$, $k = 1,2,\ldots,m_i-1$, $i = 1,\ldots,d$.
Note that the copula built on the relaxed distribution functions still satisfies Sklar's theorem for the source marginals since it produces the same realizations on the Cartesian product of the range of the source marginals, but is now uniquely defined as any marginal distribution $\tilde{F}_i(\cdot)$, $i = 1,\ldots,d$ follows an uniform distribution on $[0,1]$.

In short, we replace Step 3 in Algorithm~\ref{algo:generation} by
\begin{enumerate}
    \item[Step 3] Generate a synthetic population of vectors in $[0,1]^d$ by sampling from $C$ extended to $\tilde{F}_i(\cdot)$, $i = 1,\ldots,d$.
\end{enumerate}

Another potential issue with the proposed method is the presence of categorical random variables.
By definition, the outcomes of a categorical variable cannot be ordered.
As a consequence, such a variable has no CDF, and \eqref{eq:ecdf} cannot be built.
However, we can impose an artificial order on the outcomes by using numerical labels, over which a CDF can be described and \eqref{eq:ecdf} is now well defined.
The labeling, of course, implies a specific order that could influence the experimental outputs. Nevertheless, our numerical results suggest that the approach is robust with respect to the labeling.

\section{Numerical experiments}\label{sec:exp}
In this section, we present a series of experiments designed to assess the capability of various methods in capturing both the marginal distributions and the multivariate dependencies of a reference dataset. We place particular emphasis on techniques that incorporate copula normalization. \footnote{The source code for the methods discussed in this paper is available at \url{https://github.com/PascalJD/copulapopgen}.}

\subsection{Data sources}
For our study, we utilize data from the American Community Survey (ACS), an annual survey conducted by the U.S. Census Bureau. 
The ACS collects a variety of demographic, economic, and social information from millions of people across the U.S. 
It provides detailed data for Public Use Micro Areas (PUMAs), which are geographic units defined by the U.S. Census, each containing at least 100,000 people. 
PUMAs are unique in that they do not overlap and are nested within individual states. 
For example, the State of Maryland comprises 44 PUMAs nested within 24 counties, with each PUMA further divided into several census tracts. 
The variables extracted from the ACS, which are combined at the person level via household information, are detailed in Table~\ref{tab:data}.

In our experiments, the ACS data is treated as a fully known population. 
We create samples for training (the source dataset) and testing (the reference or target dataset) in each case.
We operate under the assumption that we have full access to both the training data and the marginal distributions of the target data.
The performance of synthetic populations generated by Bayesian Networks (BN), Tabular Variational Autoencoders (TVAE), and Conditional Tabular Generative Adversarial Networks (CTGAN), both with and without copula normalization, is compared. 
We compare our method to IPF and to an independent baseline created via bootstrapping the marginals. Additionally, we provide the empirical marginal distributions of the target population to IPF and to the methods incorporating copula normalization.
This additional data input allows these methods to potentially achieve a more accurate representation of the target population, as they are specifically designed to integrate and utilize this extended information for enhanced model fidelity.

The value of an effective population synthesizer lies not only in its ability to accurately represent a sample from a similar geographical context but also in its capacity for transferability. 
Transferability, in our context, refers to the synthesizer's ability to replicate the population characteristics of a target dataset (or geographical level) that potentially differs in marginal distributions from the source dataset.
Therefore, in this study, we explore population synthesis at three distinct geographical levels. Our first experiment is conducted at the state level, focusing on replicating the population structure of the State of Maryland (MD). 
The second set of experiments examines the spatial transferability of our models between populations at the same geographical level.
This involves training the models with data from one county and then applying the learned framework to another county. 
To further validate this approach, we conduct a similar transferability exercise between two PUMAs.
The final set of experiments investigates the transferability from a larger to a smaller geographical level. 
This includes transitioning from the county level to a PUMA and, from a PUMA to a census tract.
These experiments are particularly insightful, as they allow us to assess whether our generative models can maintain their performance when applied to populations from different geographies, which may exhibit distinct marginal distributions. 

\begin{table}[htbp]
\centering
\setlength{\tabcolsep}{0.5em} 
\renewcommand{\arraystretch}{1.5} 
\begin{tabular}{|m{0.1\textwidth}|m{0.15\textwidth}|m{0.3\textwidth}|m{0.35\textwidth}|}
\hline
\textbf{Level} & \textbf{Name} & \textbf{Description} & \textbf{Number of Values} \\
\hline
\multirow{4}{*}{Individual} & AGEP & Age of person & Age range from 0 to 99 years \\
\cline{2-4}
 & SEX & Gender of person & 2 (Male, Female) \\
\cline{2-4}
 & RAC1P & Race of person & 9 (Various races and combinations) \\
\cline{2-4}
 & ESR & Employment status & 6 (Different employment statuses) \\
\hline
\multirow{5}{*}{Household} & HINCP & Household income over the past year & 6 (Income ranges) \\
\cline{2-4}
 & HHT & Household/family type & 7 (Types of household arrangements) \\
\cline{2-4}
 & NP & Number of persons in the household & 7 (Ranges of persons in a household) \\
\cline{2-4}
 & WIF & Workers in family during the past 12 months & 4 (Number of workers) \\
\cline{2-4}
 & HUPAC & Household presence and age of children & 4 (Household with/without children and their ages) \\
\hline
\end{tabular}
\caption{Summary of demographic and socioeconomic variables from the survey data}
\label{tab:data}
\end{table}

\subsection{Evaluation metrics}
Evaluating the performance of our models involves considering two equally important aspects. 
The first one certainly is accuracy, in the sense of fitting the target data as accurately as possible. 
However, it is equally relevant to achieve this task while also maintaining the diversity of the synthetic data.

In order to compare the synthetic and reference data distributions, we calculate the Standardized Root Mean Squared Error (SRMSE), defined by \citet{SunErat15} as:
\begin{equation*}
\text{SRMSE} = \sqrt{M\sum_{m_1=1}^{M_1}\ldots\sum_{m_d=1}^{M_d}\left(\pi_{m_1\ldots m_d} - \hat{\pi}_{m_1\ldots m_d}\right)^2},
\label{eq:SRMSE}
\end{equation*}
where \( d \) is the number of variables, \( M_i \) denotes the number of categories for variable \( i \), \( \pi_{m_1 \ldots m_d} \) and \( \hat{\pi}_{m_1\ldots m_d} \) represent the relative frequencies of a particular combination in the reference data and in the synthetic data, respectively, and \( M = \prod_{i=1}^d M_i \).
SRMSE captures whether a combination of synthetic data appears in the actual data in similar proportions. 
A SRMSE value of $0$ indicates a perfect match, whereas larger values signify a growing disparity between the reference and synthetic data.
We follow the approach of \citet{BoryRichPere19}, and project the SRMSE over subsets of $n$ variables, with $n$ ranging from~$1$~to~$5$.
For $n = 1$, the SRMSE evaluates the fit of the marginals, whereas for $n > 1$, it assesses the fit of the multivariate dependencies.

On the other hand, we make use of the so-called sampled zeros (SZ) to evaluate the diversity of the synthetic data. SZ is the count of the generated combinations of variables that exist in the reference data but are unobserved in the training set~\citep{GarrBoryPereRich20, kim2023deep}. Lower SZ indicates lower diversity. In particular, a SZ count of $0$ implies that the model failed to produce unseen realistic examples, while larger SZ values indicate the model's capability to generate out-of-sample examples. 
{Similarly, we consider structural zeros (STZ), which are infeasible combinations of attributes that do not exist in the population. 
Lower counts of STZ suggest higher feasibility of the generated data, reflecting its quality and usefulness. 
To quantitatively assess the generated data, we measure precision, recall, and the F1 score. 
Precision reflects the proportion of generated data that are feasible, recall measures how well the synthetic data covers the attribute combinations present in the population, and the F1 score harmonizes these metrics to provide a single measure of data quality~\citep{kim2023deep}.

\subsection{Population synthesis at the state level}\label{subsection:state}
In this first experiment, we design a population synthesis task that aims at reproducing the data structure of the State of Maryland.
To this end, we allocate 80\% of Maryland's dataset to training and the remaining 20\% to testing.
The training set for this experiment comprises 282,009 data points.
In this scenario, the training data is not only representative of the population but it also maintains similar marginal distributions to the reference data.

Table~\ref{tab:state} details the performance, in terms of SRMSE and SZ, of the conventional tabular generative models under consideration: CTGAN, TVAE, and BN. 
We assess these models both with and without the integration of the proposed copula normalization. 
In addition, we evaluate the performance of IPF and an independent baseline.
\footnote{For IPF, we use the implementation provided in the R package developed by \citet{Ward2020}.}

The independent baseline involves generating a synthetic population by independently sampling each variable from its marginal distribution observed in the training population. 
Decoupling the dependencies between variables in this manner,  an upper bound on the SRMSE in high dimensions dimensions is stablished.
This is because the lack of dependencies results in a simplified, albeit unrealistic, population structure.
Still, the independent sampling process can create sampling zeros because, when variables are sampled independently, the resulting combinations may include attribute pairs that were not observed together in the training data, thus representing potential sampling zeros.

The similar SRMSE values observed in Table \ref{tab:state} for models with and without copula normalization suggest that the multivariate dependency is effectively preserved.
This success is primarily attributed to the underlying generative model that learns the copula, as described in step 3 of Algorithm \ref{algo:generation}.

\begin{table}[htbp]
\centering
\begin{tabular}{|l|c|c|c|c|c|}
\hline
\textbf{Method}& \textbf{SRMSE 1} & \textbf{SRMSE 2} & \textbf{SRMSE 3} & \textbf{SRMSE 4} & \textbf{SRMSE 5} \\ \hline
Independent            & 0.0147           & 0.4258           & 1.2229           & 2.6690           & 5.2975   \\ \hline
IPF            & 0.0092           & 0.1097           & 0.3918           & 1.1642           & 3.2045           \\ \hline
CTGAN          & 0.0707           & 0.2705           & 0.6819           & 1.5117           & 3.2031           \\ \hline
CTGAN Copula   & 0.0683           & 0.2647           & 0.6698           & 1.4871           & 3.1540           \\ \hline
TVAE           & 0.1652           & 0.4464           & 0.9592           & 1.9259           & 3.7910           \\ \hline
TVAE Copula    & 0.1618           & 0.4395           & 0.9436           & 1.8937           & 3.7269           \\ \hline
BN             & 0.0166           & 0.0812           & 0.2566           & 0.6820           & 1.6808           \\ \hline
BN Copula      & 0.0095           & 0.0774           & 0.2577           & 0.6917           & 1.7041           \\ \hline
\end{tabular}
\caption{Standardized root mean squared error (SRMSE) for the state level experiment}
\label{tab:state}
\end{table}

Furthermore, when computing the structural zeros and the F1 score, one faces the challenge of not having access to all feasible attribute combinations within the target population--a dataset we naturally do not possess. 
Following the methodology established by \citet{kim2023deep}, we overcome this limitation by assuming that our entire state sample encapsulates the complete population. 
This assumption enables us to estimate structural zeros by identifying combinations absent in this comprehensive sample, thus treating it as a complete representation of feasible combinations. 
We acknowledge that this is a strong assumption, which could lead to the misclassification of sampling zeros as structural zeros because rare, but feasible, combinations, can be absent from the dataset, due to their low occurrence probabilities.
As a result, the number of structural zeros is typically overestimated.

Prior to computing the zeros, we remove the age variable.
The age variable can take on a wide range of values, potentially leading to a sparse distribution across many age-specific subgroups in the data. 
By removing the age variable, we reduce the occurrence of false zeros, zeros that appear because there is not enough data in specific age brackets.
The F1 score is then calculated using this framework to evaluate the balance between; precision, reflecting the proportion of feasible synthetic data; and recall, which measures the coverage of these feasible combinations within the generated data.
These metrics are presented in Table~\ref{tab:zeros}.
IPF shows notable limitations in terms of sampling zeros, recall, and F1 score, as it fails to generalize to new samples--a shortcoming not observed in the rest of the methods.

\begin{table}[ht]
\centering
\begin{tabular}{|l|c|c|c|c|c|c|}
\hline
\textbf{Method} & \textbf{Sampling Zeros} & \textbf{Structural Zeros} & \textbf{Precision} & \textbf{Recall} & \textbf{F1 Score} \\
\hline
Independent & 200  & 21289 & 0.2306 & 0.3148 & 0.2662 \\ \hline
IPF         & 0    & 0     & 1.0000 & 0.0889 & 0.1633 \\ \hline
CTGAN       & 189  & 3133  & 0.6999 & 0.3606 & 0.4759 \\ \hline
CTGAN Copula & 199  & 3502  & 0.6787 & 0.3651 & 0.4748 \\ \hline
TVAE        & 258  & 3906  & 0.6730 & 0.3967 & 0.4991 \\ \hline
TVAE Copula & 263  & 4338  & 0.6508 & 0.3990 & 0.4947 \\ \hline
BN          & 245  & 2367  & 0.7941 & 0.4505 & 0.5748 \\ \hline
BN Copula   & 236  & 2871  & 0.7583 & 0.4446 & 0.5605 \\ \hline
\end{tabular}
\caption{Diversity and feasibility of the synthesised population at the state level}
\label{tab:zeros}
\end{table}

In this experiment, we did not anticipate significant improvements in matching the marginal distributions of the target because the training distribution is representative of the target. 
Nevertheless, our results demonstrate that our method with copula normalization effectively retains the SRMSE and diversity properties of the underlying generative model. 
When the source and target populations share similar marginal distributions, both conventional generative models and copula normalization prove effective.

As we proceed with subsequent experiments, we aim to demonstrate how copula normalization excels in scenarios that involve transferability, particularly when there is a divergence in marginal distributions between the training and reference data.

\subsection{Spatial transferability at the same geographical level}\label{subsection:samelevel}

In our previous experiment, we successfully demonstrated the versatility of copula normalization across various generative models in a standard population synthesis task.
Moving forward, we illustrate the advantages of the proposed copula normalization within model transferability frameworks.
Given the small size and dimensionality of our training sets, we have excluded performance metrics for CTGAN and TVAE from this experiment. Consistent with \citet{SunErat15}, our previous results at the state level suggest that Bayesian Networks are better suited for population synthesis tasks with limited training data. This, in this section, we analyze the transferability of models between different regions within the same geographical scale. Specifically, in one task we examine the transferability of models between two counties; in another, between two PUMAs.
Our hypothesis posits that despite these regions having dissimilar marginals, the underlying population distributions share a copula (as defined in Section~\ref{def:shared_copula}).
This conjecture stems from the notion that while the marginal behaviors of different populations may vary, the underlying dependencies among the variables could remain consistent across these geographical entities.
As stated above, the training sets for the county and PUMA experiments comprise respectively 6,360 and 6,533 data points.

Table \ref{tab:samelevel} details the performance of an independent baseline, IPF, and BN; both with and without copula normalization for the latter. 
The integration of copula normalization consistently improve BN's performance in terms of SRMSE for both levels of aggregation, though its impact lessens with increasing dimensions.
This suggests that generative models benefit from incorporating empirical marginal information via copula normalization, particularly for achieving more precise matches in marginal distribution.
Copula normalization consistently outperformed IPF in terms of SRMSE, especially in the PUMA experiment where training data was limited.

\begin{table}[htbp]
\centering
\begin{tabular}{|l|l|c|c|c|c|c|}
\hline
\multicolumn{2}{|c|}{\textbf{Method}} & \textbf{SRMSE 1} & \textbf{SRMSE 2} & \textbf{SRMSE 3} & \textbf{SRMSE 4} & \textbf{SRMSE 5}  \\ \hline
\multirow{4}{*}{County} & Independent & 0.3276 & 0.8882 & 1.9079 & 3.7393 & 7.0592 \\ \cline{2-7} 
                        & IPF         & 0.0579 & 0.2890 & 0.8860 & 2.4127 & 6.2764 \\ \cline{2-7} 
                        & BN          & 0.3242 & 0.7546 & 1.5187 & 2.9677 & 5.7906 \\ \cline{2-7} 
                        & BN Copula   & 0.0205 & 0.2680 & 0.8295 & 2.0256 & 4.5293 \\ \hline\hline
\multirow{4}{*}{PUMA}   & Independent & 0.6077 & 1.3081 & 2.3829 & 4.2759 & 7.9604 \\ \cline{2-7} 
                        & IPF         & 0.2451 & 0.6776 & 1.6260 & 3.8466 & 9.2198 \\ \cline{2-7} 
                        & BN          & 0.6021 & 1.2343 & 2.2221 & 4.0211 & 7.5800 \\ \cline{2-7} 
                        & BN Copula   & 0.0289 & 0.5549 & 1.4802 & 3.2852 & 6.9759 \\ \hline
\end{tabular}
\caption{Standardized root mean squared error (SRMSE) for the spatial transferability experiment at the county and PUMA levels}
\label{tab:samelevel}
\end{table}

In this vein, Figure \ref{fig:puma} illustrates the marginal distributions for the reference, training, and synthetic populations in the PUMA-to-PUMA experiment. 
On the one hand, a noticeable divergence between the marginal distributions of the training and target populations can be observed. Interestingly, IPF appears to align better with the reference marginal distributions when the marginals of the reference and training sets are similar, as exemplified by the RAC1P variable.
On the other hand, the figure also demonstrates that BN with copula normalization more closely matches the reference data's marginal distributions across all variables than IPF, highlighting the effectiveness of copula normalization in ensuring fidelity to the target population's distribution characteristics.
\begin{figure}[htbp]
    \centering
    \includegraphics[width=1\textwidth]{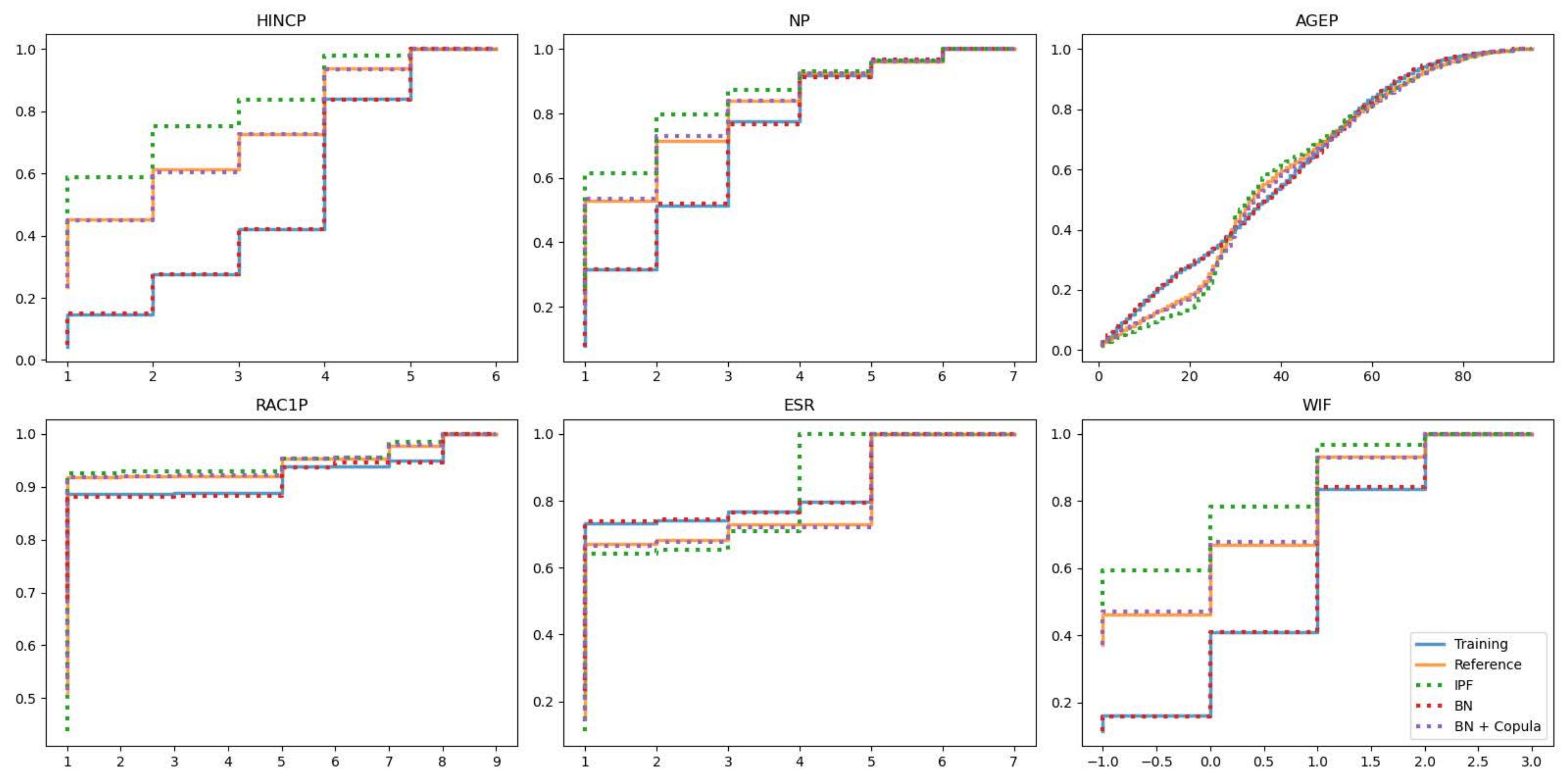}
    \caption{Marginals fit for the spatial transferability experiment at the PUMA level}
    \label{fig:puma} 
\end{figure}

As discussed in Section~\ref{def:shared_copula}, we have added numerical labels to the outcomes of the categorical variables (HHT, SEX, ESR, and RAC1P) in order to use the ECDF~\eqref{eq:ecdf}.
To explore the impact of the label ordering, 
particularly in the context of our PUMA-to-PUMA transfer experiment, 
we conducted the experiment with 100 random permutations of the categorical variables' outcomes. 
Our objective here is to ascertain whether the SRMSE values for BN with copula normalization remain consistent regardless of the ordering. 
The results of this analysis are presented in Table \ref{tab:categorical}, which illustrates the mean and standard deviation of the SRMSE values across these permutations, providing evidence that our framework's performance is robust to the ordering of categorical variables.

\begin{table}[h]
\centering
\begin{tabular}{|c|c|c|c|c|c|}
\hline
\textbf{Method} & \textbf{SRMSE 1} & \textbf{SRMSE 2} & \textbf{SRMSE 3} & \textbf{SRMSE 4} & \textbf{SRMSE 5}\\
\hline
BN Copula & 0.0333 $\pm$ 0.0032 & 0.5542 $\pm$ 0.0123 & 1.4752 $\pm$ 0.0375 & 3.2666 $\pm$ 0.0846 & 6.9238 $\pm$ 0.1651 \\ \hline
\end{tabular}
\caption{SRMSE results for the BN with copula normalization across 100 random permutations of categorical variables\mfabian{' outcomes} in the PUMA-to-PUMA experiment}
\label{tab:categorical}
\end{table}

\subsection{Spatial transferability at smaller geographical level}

In this final numerical experiment, we investigate the transferability of models from larger to smaller geographical levels using data from the ACS. 
We establish two scenarios for evaluation. 
In the first, the model is trained on data at the county level and then transferred to a PUMA within that county. 
In the second scenario, the training data is sourced from a PUMA, and the reference data is from a census tract within that PUMA. 
As stated above, the training set for the county to PUMA transferability experiment comprises 36,954 data points and the one for the PUMA to census tract comprises 6,533 data points.

Analogously to the previous sections, Table \ref{tab:countypuma} presents the results for the transfer from the county level to PUMA level.
We compare the performance of an independent baseline, IPF, and BN with and without copula normalization.
Notably, IPF underperforms compared to the independent baseline, suggesting that, when the sample size is small, population heterogeneity may not be sufficient to accurately calibrate a contingency table.
Hence, the marginal constraints imposed by IPF can hinder, rather than enhance, the fitting of the joint distribution.
This is a significant result aligned with the findings of  \citet{SunErat15}.

On the other hand, a substantial advantage can be observed when implementing copula normalization in BN, as it yields markedly lower SRMSE compared to its counterpart. 
Notably, the SRMSE values for the BN with copula normalization are an order of magnitude lower when just one dimension is analyzed, subsequently maintaining a consistent advantage as dimensionality increases. 
This enhancement indicates that copula normalization effectively captures the marginal distributions, leading to a more accurate representation of the reference population; certainly a desirable capability in these cases of small geographic areas, where traditional methods seem to struggle due to limited data variability and size.

\begin{table}[htbp]
\centering
\begin{tabular}{|l|c|c|c|c|c|}
\hline
\textbf{Method} & \textbf{SRMSE 1} & \textbf{SRMSE 2} & \textbf{SRMSE 3} & \textbf{SRMSE 4} & \textbf{SRMSE 5} \\ \hline
Independent & 0.3442 & 0.8723 & 1.6720 & 3.0777 & 5.9539 \\\hline
IPF         & 0.7020 & 1.5510 & 3.0708 & 6.2470 & 13.4378 \\\hline
BN          & 0.3334 & 0.7402 & 1.4119 & 2.6954 & 5.4334 \\\hline
BN Copula   & 0.0350 & 0.3221 & 0.8658 & 1.9979 & 4.5310 \\\hline
\end{tabular}
\caption{Standardized root mean squared error (SRMSE) for the spatial transferability experiment from the county to PUMA levels}
\label{tab:countypuma}
\end{table}

In our final experiment, we demonstrate the efficacy of copula normalization in facilitating synthetic data generation at smaller geographical levels, such as a census tract, where granular data is typically unavailable. 
This experiment focuses on model transferability from a higher level (PUMA) to a lower level (census tract).

It is worth noting at this point that, although the ACS data provides detailed information at the PUMA level, it this granularity does not extend to the census tract level. 
To address this gap, we utilize marginal counts from the U.S. decennial Census Data, enabling us to provide marginal information of the reference to IPF and copula normalization for RAC1P, NP, HINCP, HUPAC, and AGEP variables. 
Given the absence of granular data at the census tract level, our evaluation relies on a visual assessment of the fit of the marginal distributions. Figure \ref{fig:ctract}, analogous to Figure \ref{fig:puma} in the previous section, offers a graphical comparison of the marginal distributions produced by IPF, BN, and BN with copula normalization. 
This visual representation showcases how each method replicates the observed marginal distributions within the census tracts.
The alignment of the synthetic populations generated by BN with copula normalization with the actual marginal distributions is clear in the closeness of the graph lines. 
This consistency underscores the utility and effectiveness of copula normalization in scenarios characterized by limited detailed data. 
It highlights the method’s potential in enhancing model transferability, particularly in synthesizing populations for smaller geographic areas like census tracts, where data scarcity often poses a significant challenge.

\begin{figure}[htbp]
    \centering
    \includegraphics[width=1\textwidth]{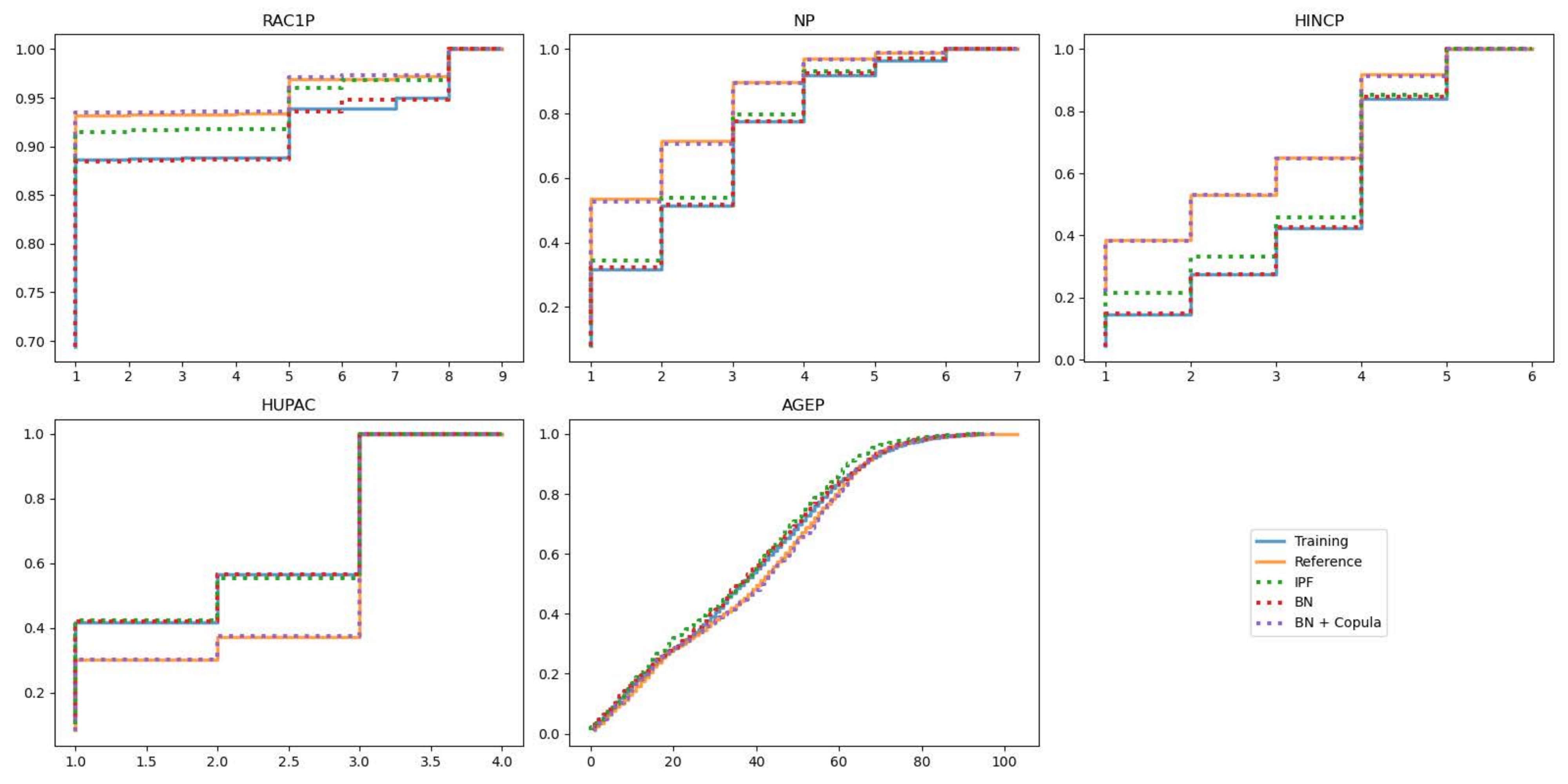}
    \caption{Marginals fit for the spatial transferability experiment at the census tract level}
    \label{fig:ctract} 
\end{figure}

\section{Conclusion}\label{sec:conclusion}
We introduce in this paper a framework that combines copula theory with traditional machine learning generative methods to decouple the learning of the dependency structure among variables from the peculiarities of their marginal distributions. 
Central to our method is the concept of model transferability, which refers to the ability of a model to be effectively applied to different but related datasets or environments.
Our approach allows to leverage insights from data-rich environments and applying them to contexts with limited data availability.
More precisely, our method involves normalizing the source population using empirical cumulative distribution functions (ECDFs), thereby treating observations as realizations of an underlying copula. 
A generative model, selected based on the specific needs of the task, is employed to learn this copula and to sample a normalized synthetic population. 
The normalized generated data is then transformed using the pseudo-inverses of the reference population’s cumulative distribution functions, effectively mapping it back to the original sample space and injecting the marginal information of the reference. 
To address the challenges associated with discrete distributions, we also introduce a heuristic approach that relaxes discrete marginals into continuous distributions, considering linear interpolations between consecutive values of their ECDFs. 
This strategy allows for the generation of synthetic populations that accurately reflect a reference population, potentially comprising a mix of continuous and discrete variables known only through its marginal distributions and a sample from a similar but distinct population.
Our copula normalization approach is designed to be model-agnostic, offering a simple yet effective methodology that can be applied across different models without the need for tailoring to specific model architectures.

Recognizing the critical role of geographical scales in population synthesis, we conducted a series of experiments at various levels, including state, county, PUMA, and census tract. 
Specifically, we focused on the transferability of models between different regions, both within the same scale (e.g., between counties and between PUMAs) and across scales (from county to PUMA and PUMA to census tract).
To assess the effectiveness of these methods, we utilized standardized root mean squared error (SRMSE) for accuracy, sampled zeros for diversity, and structural zeros for feasibility.
Our copula framework was applied with several machine learning generative methods: Bayesian Network, Conditional Tabular GAN, and Tabular VAE.
We compared its performance against those of Iterative Proportional Fitting (IPF) and a baseline population derived from bootstrapping the source sample. 
The empirical results across these geographical levels demonstrate that our copula framework improves the performance of all evaluated machine learning methods in matching the marginals of the reference data, and consistently outperforms IPF in transferability experiments in terms of SRMSE, while also naturally introducing unique observations not present in the original sample.
In particular, BN in combination with our copula framework achieved the overall best performance.
Nevertheless, our focus on discrete variables leaves the performance of the copula framework with continuous variables an open question for future research. 
Evaluating copula normalization across various contexts, including large-scale continuous datasets where deep learning models integrated with copulas might offer superior performance compared to BN, also poses an interesting avenue for future research.
Furthermore, this work opens the door to leveraging data from multiple sources, such as combining disaggregated survey data with aggregated administrative data. This approach is particularly vital in addressing the challenges of limited sample sizes in small areas, enabling the generation of precise population statistics and travel indicator estimates at granular geographical levels. These comprehensive insights are crucial for policy analysis and decision-making in governmental agencies.

\section*{Acknowledgments}
The work of Fabian Bastin is supported by the Natural Sciences and Engineering Research Council of Canada [Discovery Grant 2022-04400]. 
Pascal Jutras-Dubé was furthermore supported by a graduate grant under the NSERC CREATE Program on Machine Learning in Quantitative Finance and Business Analytics~(Fin-ML)

\bibliography{references}
\bibliographystyle{abbrvnat}

\end{document}